\documentclass{ecai} 

\usepackage{amsmath}
\usepackage{amssymb}
\usepackage{algorithm}
\usepackage{algorithmic}

\usepackage{latexsym}
\usepackage{amssymb}
\usepackage{amsmath}
\usepackage{amsthm}
\usepackage{booktabs}
\usepackage{enumitem}
\usepackage{graphicx}
\usepackage{color}
\usepackage{xurl}
\usepackage{tabularx}
\usepackage{subfigure}
\usepackage{ifthen}  % needed for \ifthenelse
\newcommand{\BibTeX}{B\kern-.05em{\sc i\kern-.025em b}\kern-.08em\TeX}

\begin{document}

%%%%%%%%%%%%%%%%%%%%%%%%%%%%%%%%%%%%%%%%%%%%%%%%%%%%%%%%%%%%%%%%%%%%%%%%

\begin{frontmatter}

%%% Use this command to specify your submission number.
%%% In doubleblind mode, it will be printed on the first page.

\paperid{6113} 
\def\mode{1}  % Change this to 0, 1, or 2 as needed
%%% Use this command to specify the title of your paper.
\newboolean{showmain}
\newboolean{showsupp}

\ifthenelse{\equal{\mode}{0}}{
  \setboolean{showmain}{true}
  \setboolean{showsupp}{true}
}{
  \ifthenelse{\equal{\mode}{1}}{
    \setboolean{showmain}{true}
    \setboolean{showsupp}{false}
  }{
    \setboolean{showmain}{false}
    \setboolean{showsupp}{true}
  }
}
\title{Owen Sampling Accelerates Contribution\\Estimation in Federated Learning}

\author[A]{\fnms{Hossein}~\snm{KhademSohi}}
\author[A,B]{\fnms{Hadi}~\snm{Hemmati}}
\author[C]{\fnms{Jiayu}~\snm{Zhou}}
\author[A]{\fnms{Steve}~\snm{Drew}\thanks{Corresponding Author. Email: steve.drew@ucalgary.ca.}}

\address[A]{Department of Electrical and Software Engineering, University of Calgary, Calgary, Alberta, Canada}
\address[B]{Department of Electrical Engineering and Computer Science, York University, Toronto, Ontario, Canada}
\address[C]{School of Information, University of Michigan, Ann Arbor, Michigan, USA}

%%% Use this environment to include an abstract of your paper.

\begin{abstract}
Federated Learning (FL) aggregates information from multiple clients to train a shared global model without exposing raw data. Accurately estimating each client’s contribution is essential not just for fair rewards, but for selecting the most useful clients so the global model converges faster. The Shapley value is the principled choice for this, yet exact computation scales exponentially with the number of clients, making it infeasible for real-world FL deployments with many participants. In this paper, we propose \emph{FedOwen}, an efficient federated contribution evaluation framework adopting Owen sampling to approximate Shapley values under the same total evaluation budget as the existing methods, while keeping the approximation error below a small threshold. In addition, FedOwen applies an adaptive client selection strategy that balances exploiting high-value clients with exploring under-sampled ones, avoiding bias toward a narrow subset, and uncovering rare but informative data. Under a fixed valuation cost, FedOwen achieves up to 23\% improvement in final model accuracy within the same number of communication rounds, compared to state-of-the-art baselines on non-IID benchmarks.
\noindent\textbf{Code:} \url{https://github.com/hoseinkhs/AdaptiveSelectionFL}
\end{abstract}

\end{frontmatter}

%%%%%%%%%%%%%%%%%%%%%%%%%%%%%%%%%%%%%%%%%%%%%%%%%%%%%%%%%%%%%%%%%%%%%%%%

\section{Introduction}

Federated Learning (FL) enables collaborative training over decentralized data without sharing raw data, which is essential in privacy-sensitive domains such as healthcare \cite{antunes2022federated}, finance \cite{suzumura2022federated}, and telecommunications \cite{mangla2022application}. In practice, many clients may be available and the server selects a subset each round. To make fast progress and support incentives, it needs fair and informative estimates of each client’s contribution \cite{fu2023client}.

The Shapley value \cite{winter2002shapley} provides a principled way to attribute contributions by averaging marginal effects over all coalitions. However, exact computation scales exponentially with the number of clients, which makes real-time or large-scale use impractical.

Approximate methods reduce complexity but typically rely on a \textit{valuation dataset} \cite{guo2024contribution, shyn2021fedccea}. Because every sampled coalition must be evaluated on the full set, cost grows with both dataset size and the number of samples. A single dataset can also misrepresent utility, and selecting clients solely by these scores can create a positive-feedback bias toward already high-scoring clients \cite{huang2021shapley}.

To address these limits, we propose \emph{FedOwen}, a contribution evaluation framework for FL that adopts Owen sampling \cite{owen1972multilinear,okhrati2021multilinear}, a binomial antithetic-pair estimator that preserves the unbiased nature of standard Monte Carlo while running more efficiently in large-scale FL. We also use a \emph{truncated} version that stops evaluating a permutation once the remaining possible gain drops below a small threshold~\(\eta\). This early stop removes many model evaluations without hurting accuracy. In addition to this estimator, we develop an adaptive client-selection rule that balances exploration and exploitation from round to round, further enhancing the efficiency of our sampling. This selection mechanism is inspired by the ideas in contribution-based selection~\cite{lin2022contribution} and adding a penalizer to reduce bias. The rule keeps the pool diverse and participation balanced, especially important for long-tailed data and non-IID splits~\cite{zhang2022data}. Our approach models the client selection as a Multi-Armed Bandit problem with an $\epsilon$-greedy exploration combined with contribution-based client ranking. By alternating between random exploration and selecting the top clients, it achieves higher accuracy more quickly and remains robust in diverse FL settings.
The source code and all experimental scripts are available in the github repo.
We summarise our contributions as follows:

%\vspace{-4pt}

\begin{itemize}
  \item We propose FedOwen, a novel contribution evaluation method in FL with Owen sampling, enhanced by cost‑aware $\eta$‑truncation that lowers valuation-set cost while maintaining unbiased Shapley estimates. To our knowledge, we are the first to bring Owen's antithetic estimator in FL, enabling the application of federated contribution evaluation in large-scale client pools.

  \item We develop an $\epsilon$-greedy client selection rule that alternates between high‑value and under‑explored clients, preventing ``rich‑get‑richer'' bias and accelerating convergence.

  \item We conduct an empirical study under a strict budget‑parity protocol where all methods are limited to the same valuation calls per round. Our approach outperforms five recent baselines on five datasets and six heterogeneity settings, achieving up to 23\% higher final accuracy and therefore reaching target accuracy in fewer communication rounds.
\end{itemize}

While we do use a small server-held evaluation set, our cost reduction stems from reducing the \emph{number of utility evaluations} via early truncation rather than avoiding evaluation data altogether.

\section{Related Work}

Estimating client contributions in FL is crucial for fair resource allocation and incentivizing continued participation \cite{yu2020fairness}. Existing methods often employ concepts such as the Shapley value, which can be fair but computationally expensive. Recent techniques seek to improve communication efficiency, accuracy, robustness, and incentive mechanisms under practical resource and security constraints.

\subsection{Communication and Computation Efficiency}
Early Shapley-based FL works optimize communication/computation \cite{wang2020principled,tang2021optimizing,fan2022fair}, while GTG-Shapley \cite{liu2022gtg} reconstructs models from gradients to cut cost under non-IID data. Recent systems adapt client counts \cite{li2024adafl}, weight aggregation by contribution \cite{du2025hfedcwa}, or use RL/combinatorial scheduling \cite{pan2025rfcsc,li2025optimal}. These improve efficiency but often rely on aggressive approximations or assumptions that can distort valuations at scale.

CTFL \cite{wang2024fast} proposes a fast and interpretable rule-based approach to estimate client contributions, significantly reducing computation times. FLContrib \cite{ghosh2024don} filters malicious influences and balances client contributions via a history-aware technique. GREEDYFED \cite{singhal2024greedy} uses a biased, fast Shapley approximation to pick top-contributing clients under tight resource constraints.
These approaches significantly reduce overhead but commonly rely on aggressive approximations or simplified assumptions, which can lead to less precise contribution valuations and limit their scalability in very large federations.
\subsection{Valuation Precision Algorithms}

Data Shapley \cite{ghorbani2019data} is introduced in supervised learning to quantify the importance of each data point, though calculating exact Shapley values can be expensive. Pairwise Correlated Agreement (PCA) \cite{lv2021data} avoids needing a test set, improving accuracy while motivating honest contributions. SPACE \cite{chen2024space} offers higher speed and accuracy in estimating individual effects. Meanwhile, techniques like \cite{yang2024maverick} modify Shapley value computations to handle real-world constraints and client-specific conditions for fairer rewards.

FedGAC \cite{yu2025fedgac} focuses on critical learning periods for strategic client selection, ensuring better generalization.  WeightedSHAP \cite{kwon2022weightedshap} adapts Shapley values by learning different weights from data, addressing issues with uniform weighting. The Banzhaf value \cite{wang2023data} provides a robust alternative in noisy settings, improving safety margins and computational efficiency.  
ShapFed-WA \cite{tastan2024redefining} adds a Shapley score for every class and uses those scores to weight the aggregation. The extra detail is helpful, but keeping track of so many numbers can slow down deep models.

Many estimators trade unbiasedness for speed or use ad-hoc weights, risking misses on rare classes. Beyond permutation sampling, variance/consistency trade-offs are documented \cite{mitchell2022sampling}; marginal-free estimators \cite{kolpaczki2024approximating} can be competitive offline but drop axioms we require for bandit-driven selection and aggregation. We focus on an unbiased, budget-parity estimator that plugs into $\epsilon$-greedy selection; a broader shapiq-based benchmark \cite{muschalik2024shapiq} is left to future work.

\subsection{Robustness and Security}

Shapley Value-Based Client Selection \cite{zheng2022secure} (S-FedAvg) filters out irrelevant or malicious data to protect the central model from corruption. ShapleyFL \cite{sun2023shapleyfl} detects which clients have critical data and fixes corrupted samples. A fuzzy Shapley value mechanism \cite{yang2024federated} adds fuzzy logic to handle uncertainties in participation, ensuring Pareto-optimal payouts. 

Measure Contribution of Participants in Federated Learning \cite{wang2019measure} deals with horizontal and vertical FL by either deleting instances or using Shapley values to reveal feature significance. WTDP-Shapley \cite{yang2022wtdp} refines Shapley values for incentive-based selection under practical FL constraints, especially for safety inspections. Game of Gradients \cite{nagalapatti2021game} uses Shapley value ideas to stabilize model performance against irrelevant or low-quality data.

New methods also integrate security solutions. For instance, \cite{jalali2025federated} leverages blockchain in edge computing to secure and fairly organize clients. A Stackelberg game method for power and client selection in IoMT is presented in \cite{liu2025game}, while \cite{chen2025pifl} merges blockchain and FL to secure IoT edge devices.

Most defenses target single attack types and treat contribution scores as fixed, leaving gaps against adaptive adversaries. Continually updated, diversity-preserving selection remains necessary.

\subsection{Data-Free Contribution Estimation}

FedCM \cite{xu2021fedcm} relies on attention mechanisms to measure each client’s influence across rounds in real time, enhancing robustness and accuracy. FRECA \cite{zhang2024towards} operates solely on local updates (based on FedTruth \cite{ebron2023fedtruth}), avoiding external validation datasets and mitigating hostile attacks.

Data-free proxies (e.g., attention or gradient alignment) avoid server-side validation but can misrepresent true impact under non-IID or strategic behavior. Balancing robustness with faithful utility measurement remains open.

\subsection{Incentive- and Bandit-Driven Client Selection}

FairFed \cite{liu2024fairfed} applies cooperative Shapley values with dynamic adaptation to participant behavior changes. A gradient-based Shapley approximation strategy for profit sharing is proposed in \cite{song2019profit}. FLamma \cite{javaherian2025incentive} employs a Stackelberg game framework to manage contributions and rewards, while \cite{yang2022federated} uses an analytic hierarchy process (AHP) to include data quality and costs in Shapley value–based incentives.

\paragraph{Multi-Armed Bandits for Client Scheduling}
Bandit-style schedulers \cite{shi2021federated,xia2020multi,wu2023fedab,chen2024federated} improve convergence but may bias toward immediate high contributors. Our $\eta$-truncated Owen sampler with an $\epsilon$-greedy hybrid selector aims to trim valuation cost while maintaining diversity and fairness.

These strategies often assume predictable client behavior and reward patterns. As a result, they tend to favor clients with immediate high contributions, which can bias participation and weaken long-term fairness and robustness when conditions change.

Overall, research on FL contribution estimation blends game-theoretic guarantees (Shapley, Owen) with practical mechanisms (for example, gradient compression, blockchain, and bandit scheduling). Our approach follows this line: an $\eta$-truncated Owen sampler together with an $\epsilon$-greedy hybrid selector reduces valuation cost without sacrificing resolution, improving scalability and fairness.

\section{Methodology}

Our aim is to assign each client a fair \emph{contribution score} $\phi_i$.
To measure how helpful a client really is, the server keeps a small validation set $\mathcal D_{\text{val}}$ that never leaves the server.
After every communication round the server scores each returned model $W_i^{(t+1)}$ on $\mathcal D_{\text{val}}$ with a fixed metric (such as accuracy). Those validation scores act as the utility function for any group of clients, and they feed straight into our game-theoretic machinery.
FedOwen uses only the server-held evaluation set and model weights already exchanged in vanilla FL; no additional client-side data or metadata is transmitted.
The rest of this section explains how we turn these utilities into the Shapley-based scores and then use those scores in FedOwen for selection and aggregation.
\subsection{Background}

Game theory-based methods are well-regarded for their fairness in contribution evaluation. Among these, the Shapley value is widely recognized for its fairness properties, such as symmetry and additivity. In the FL context, each client is considered a player, and the set $N$ of all clients is equivalent to the set $X$ in classical game theory formulations. The Shapley value considers all possible coalitions to fairly distribute the overall gain based on the marginal contributions of each member.
Define $N$ as the set of all clients, $S$ as a subset of $N$ excluding client $i$. Denote the value of the coalition $S$ by $v(S)$. This value is typically the performance (e.g., accuracy) of the model trained with clients in $S$.
The Shapley value for a client $i$, denoted by $\phi_i$, is then defined as follows:

\begin{equation}
\phi_i = \sum_{S \subseteq N \setminus \{i\}} \frac{|S|!(|N| - |S| - 1)!}{|N|!} \left( v(S \cup \{i\}) - v(S) \right)
\end{equation}

The Shapley value satisfies the following properties:

\begin{itemize}
    \item \textbf{Efficiency.} The total value generated by the coalition of all players is fully distributed among them:
    \begin{equation}
    \sum_{i \in N} \phi_i(v) = v(N)
    \end{equation}

    \item \textbf{Symmetry.} Players contributing equally to all coalitions receive equal value:
    \begin{equation}
    \text{If } v(S \cup \{i\}) = v(S \cup \{j\}), \forall S, \text{ then } \phi_i(v) = \phi_j(v)
    \end{equation}

    \item \textbf{Dummy Player.} If a player does not change the value of any coalition:
    \begin{equation}
    \text{If } v(S \cup \{i\}) = v(S), \forall S, \text{ then } \phi_i(v) = 0
    \end{equation}

    \item \textbf{Additivity.} For any two value functions $v$ and $w$:
    \begin{equation}
    \phi_i(v + w) = \phi_i(v) + \phi_i(w)
    \end{equation}
\end{itemize}

However, computing exact Shapley values requires evaluating $2^n$ coalitions, which is infeasible for large $n$. Thus, efficient approximation methods are necessary.

To approximate the Shapley value, Monte Carlo (MC) sampling methods \cite{castro2009polynomial} are commonly used. When combined with MC, Shapley approximates each $\phi_i$ by averaging marginal contributions across randomly sampled client permutations.

Although MC methods reduce computational demands, they still suffer from high variance and require a large number of samples for convergence.

\subsection{Owen Sampling with Early Truncation}
\label{subsec:owen_trunc}

Classical Monte‑Carlo Shapley sampling draws full client permutations, which converges slowly because every marginal gain is measured against a completely new coalition.  
Owen’s insight is to rewrite the Shapley value as an integral over an inclusion probability \(q\in[0,1]\).  
For each fixed \(q\) we generate two \textit{antithetic twins}: one coalition where a given client is present and a matching coalition where it is absent.  
Because these twins are negatively correlated, averaging their difference slashes the estimator’s variance without introducing bias.  
In FL, this lets us achieve nearly the same accuracy with fewer model evaluations. Below, we present Owen’s integral form, outline how we discretise it, and explain the early‑truncation rule used in our proposed FedOwen framework that makes it more practical.

Let $N$ be the set of clients and $n = |N|$; when an order is needed we write $X=(x_1,\dots,x_n)$.
For an inclusion probability $q\in[0,1]$, draw a Bernoulli mask $I^{(q)}=(b_1,\dots,b_n)\in\{0,1\}^n$ with $\mathbb{P}(b_j=1)=q$ independently, and let $S = I^{(q)}\odot X$ denote the induced coalition (element-wise selection).
Writing $X^{(j)}$ for the unit vector that selects client $x_j$, Owen’s multilinear extension expresses the Shapley value as
\begin{equation}
  S_j(\nu)=\int_{0}^{1} e_j(q)\,dq,
  \label{eq:shapley}
\end{equation}
where
\begin{equation}
  e_j(q)=\mathbb{E}\bigl[\nu\bigl(S\cup\{x_j\}\bigr)-\nu(S)\bigr],
\end{equation}
and the expectation is over the randomness of $I^{(q)}$.
Here $\nu(S)$ is the chosen performance metric (e.g., accuracy) of coalition $S$.
By pairing present/absent twins for each inclusion level $q$, Owen’s estimator induces negative correlation between marginals, reducing variance at a fixed number of calls to $\nu(\cdot)$ while preserving unbiasedness.

Equation~\eqref{eq:shapley} is approximated by discretising the
integral at $Q$ equally spaced inclusion probabilities
$q=i/Q\,(i=1,\dots,Q)$ and drawing $M$ independent coalitions at each
level.  For a fixed $q$ we generate the Bernoulli mask $I^{(q)}$,
visit the selected clients in a random order, and accumulate the
marginal contributions $\nu(S\cup\{x_j\})-\nu(S)$ sequentially.  During
this walk, we maintain an upper bound on the remaining potential gain.
If it falls below the tolerance~$\eta$, we terminate the walk early.
This ``truncate‑when‑nothing‑more‑can‑matter'' rule cuts the number of
model evaluations while preserving unbiasedness, so Owen sampling with
early truncation converges faster to the true Shapley value.

% \vspace{-1.5em}
\begin{figure*}[!htbp]
    \centering
    \vspace{-2.2em}
    \includegraphics[width=0.91\linewidth]{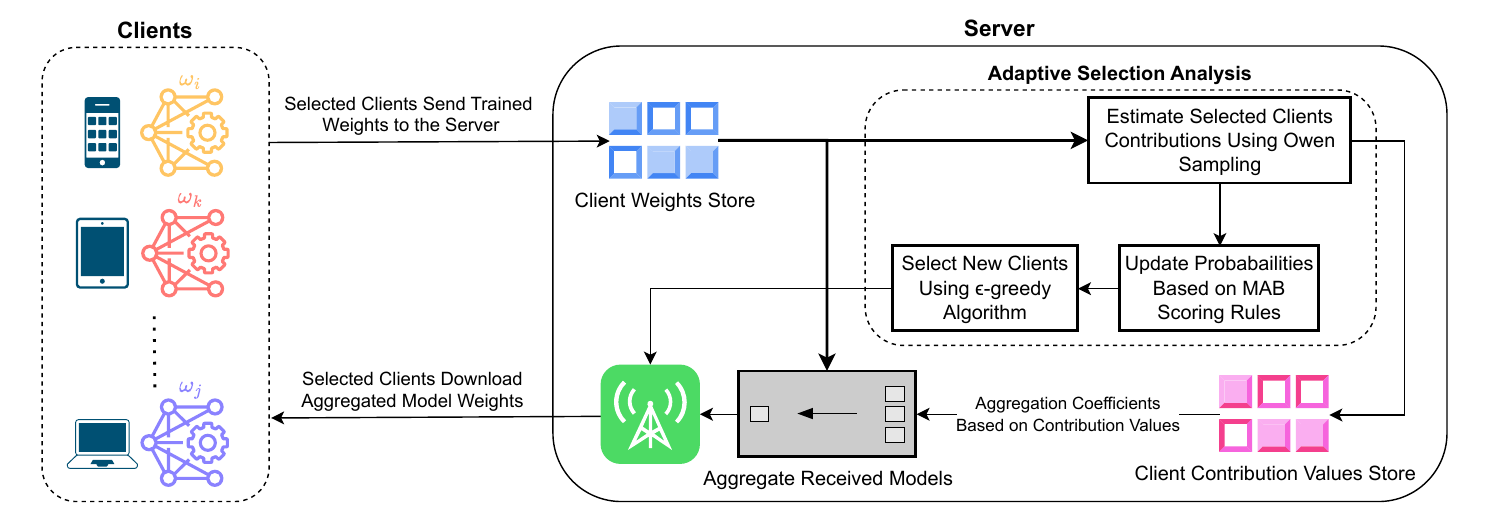}
    \vspace{-1em}
    \caption{System workflow of the proposed adaptive FL framework}
    \label{fig:system_model}
\end{figure*}
%---------------------------------------------------------------
% Algorithm : Owen Sampling with Early Truncation
%---------------------------------------------------------------
We pre-normalize the utility so that $\nu(\varnothing)=0$ and $\nu(N)=1$; the remainder $\Delta_{\text{left}}$ is thus an optimistic upper bound on the contribution still available during a walk.
\begin{algorithm}[t]
\caption{Owen Sampling with $\eta$-Truncation}
\label{alg:owen_sampling}
\begin{algorithmic}[1]
\REQUIRE Utility function $\nu$; client list $X=(x_1,\dots,x_n)$; number of inclusion levels $Q$; draws per level $M$; truncation tolerance $\eta$.
\ENSURE  Shapley estimates $\hat S=(\hat S_1,\dots,\hat S_n)$
\STATE Initialize $\hat S_j\gets 0$ for all $j$
\FOR{$q = 1/Q,\,2/Q,\,\dots,\,1$}                     \label{line:qgrid}
    \STATE Initialize level accumulator $e_j\gets 0$ for all $j$
    \FOR{$m = 1$ \textbf{to} $M$}                      \label{line:mcloop}
        \STATE Draw mask $I^{(q)}_m=(b_j)$ with $b_j\sim\mathrm{Bern}(q)$
        \STATE Let $\mathcal{P}$ be the indices with $b_j=1$ in random order
        \STATE $C\gets\varnothing$         % running coalition  (only new line)
        \STATE $\Delta_{\text{left}}\gets 1$\hfill\textit{// optimistic upper bound}
        \STATE $v_{\text{prev}}\gets 0$
        \FOR{\textbf{each} $j\in\mathcal{P}$}          \label{line:perm_walk}
            \STATE $C\gets C\cup\{x_j\}$               % grow coalition
            \STATE $h^{(q)}_{m,j}\gets \nu(C) - v_{\text{prev}}$
            \STATE $e_j \gets e_j + h^{(q)}_{m,j}$
            \STATE $\Delta_{\text{left}}\gets\Delta_{\text{left}}-h^{(q)}_{m,j}$
            \IF{$\Delta_{\text{left}}<\eta$}    \label{line:trunc}
                \STATE \textbf{break}  \hfill\textit{// truncated; add loop to preserve $N\times M$ calls}
            \ENDIF
            \STATE $v_{\text{prev}}\gets v_{\text{prev}}+h^{(q)}_{m,j}$
        \ENDFOR
    \ENDFOR
    \STATE $\hat S_j \gets \hat S_j + e_j$ for all $j$
\ENDFOR
\STATE $\hat S_j \gets \hat S_j / (Q M)$ for all $j$
\end{algorithmic}
\end{algorithm}

We describe the full Owen sampling process in Algorithm \ref{alg:owen_sampling}. Let \(X=(x_1,\dots,x_n)\) be the ordered client set and let \(X^{(j)}=\{x_j\}\) denote the singleton. The Bernoulli mask \(I^{(q)}_m\in\{0,1\}^{n}\) selects a coalition via the Hadamard selection \(I^{(q)}_m\!\odot\!X\). The random order \(\mathcal{P}\) over the selected indices ensures symmetry in the Shapley estimator. For each visited \(j\in\mathcal{P}\), with current coalition \(C\), we record the incremental utility
\(h^{(q)}_{m,j}=\nu\!\bigl(C\cup\{x_j\}\bigr)-\nu(C)\).
Summing these increments over \(M\) samples at each of the \(Q\) inclusion levels yields the unbiased estimate \(\hat S_j\).
Early truncation is performed at line~\ref{line:trunc}: once the optimistic remainder \(\Delta_{\text{left}}\) falls below \(\eta\), no subsequent client can increase the total by more than \(\eta\); the walk stops and any saved evaluations are immediately re-spent on fresh walks until the common \(N\times M\) budget is met.

%---------------------------------------------------------------
\subsection{Convergence Analysis of Owen Sampling}
\label{subsec:owen_convergence}

We show that the estimator produced by Algorithm~\ref{alg:owen_sampling}
is unbiased and converges almost surely to the exact Shapley value
$S_j(\nu)$ as the number of inclusion levels $Q$ and the per-level
sample size $M$ grow.

\paragraph{Step 1 - Integral representation.}
Recall the multilinear-extension identity \cite{owen1972multilinear}
\begin{equation}
S_j(\nu)=\int_{0}^{1} e_j(q)\,dq,
\quad
e_j(q)=\mathbb{E}\bigl[\nu\bigl((I^{(q)}\!\odot X)\cup\{x_j\}\bigr)-\nu(I^{(q)}\!\odot X)\bigr],
\label{eq:owen_integral}
\end{equation}
where $I^{(q)}\in\{0,1\}^{n}$ has i.i.d.\ $\mathrm{Bernoulli}(q)$ entries.

\paragraph{Step 2 - Monte Carlo estimate at each $q$.}
For a fixed inclusion probability $q$, the strong law of large numbers gives
\begin{equation}
\bar e_j^{(M)}(q)
:=\frac{1}{M}\sum_{m=1}^{M}
\bigl[\nu(I^{(q)}_m\!\circ X+X^{(j)})-\nu(I^{(q)}_m\!\circ X)\bigr]
\xrightarrow[M\to\infty]{\mathrm{a.s.}}
e_j(q).
\label{eq:mc_consistency}
\end{equation}

\paragraph{Step 3 - Riemann approximation of the integral.}
Since $e_j(q)$ is bounded on $[0,1]$, it is Riemann integrable. Hence
\begin{equation}
\frac{1}{Q}\sum_{i=1}^{Q}
e_j\!\bigl(\tfrac{i}{Q}\bigr)
\xrightarrow[Q\to\infty]{}\!
\int_{0}^{1}e_j(q)\,dq
= S_j(\nu).
\label{eq:riemann}
\end{equation}

\paragraph{Step 4 - Two-stage limit.}
Define
\begin{equation}
\hat S_j^{(Q,M)}(\nu)
:=\frac{1}{Q\,M}\sum_{i=1}^{Q}\sum_{m=1}^{M}
\bigl[\nu(I^{(i/Q)}_m\!\circ X+X^{(j)})
      -\nu(I^{(i/Q)}_m\!\circ X)\bigr].
\label{eq:owen_estimator}
\end{equation}
Combining \eqref{eq:mc_consistency} and \eqref{eq:riemann} and
applying the bounded convergence theorem yields
\begin{equation}
\lim_{Q \to \infty} \, \lim_{M \to \infty} \, 
\hat{S}_j^{(Q, M)}(\nu) 
= S_j(\nu) 
\quad \text{(a.s.)},
\end{equation}

\paragraph{Step 5 - Unbiasedness under early truncation.}
Algorithm~\ref{alg:owen_sampling} stops a permutation walk when the
optimistic remainder $\Delta_{\mathrm{left}}$ falls below $\eta$
(line~\ref{line:trunc}), then immediately starts a fresh walk so that
each level still uses $N\times M$ utility evaluations.  Since the
decision to truncate depends only on already‐observed increments, each
incremental gain remains mean‐preserving.  Thus the truncated
permutations yield the same expectation as full walks, and all of the
above convergence arguments remain valid.

\subsection{Adaptive Performance–Driven Client Selection}
\label{subsec:mab_selection}

\textbf{Adaptive selection.}
Owen-sampled contributions are fair and efficient, but using them alone can sideline clients whose seemingly low value hides rare, useful signals. FedOwen couples these scores with an adaptive scheduler that balances exploitation and exploration. After each valuation, the server selects $k$ clients via an $\epsilon$-greedy multi-armed bandit: with probability $\epsilon$ it explores under-sampled/low-scored clients; otherwise it exploits top contributors augmented by an optimism-in-the-face-of-uncertainty bonus. This mix preserves diversity, avoids ``rich-get-richer'' bias, and accelerates convergence—even if it occasionally departs from strictly proportional participation by score alone.

\vspace{0.4em}\noindent\textbf{Server workflow.}  
As illustrated in Fig.~\ref{fig:system_model}, the server (i) selects
clients, (ii) aggregates their updates to the global model, and (iii)
re-estimates the contribution vector
$\boldsymbol\phi=(\phi_1,\dots,\phi_n)$ for the following round.  
Clients train locally on their private data and return model updates.

\vspace{0.4em}\noindent\textbf{Exploration versus exploitation.}
Algorithm~\ref{alg:adaptive_selection} demonstrates the behaviour
implemented in the optimisation-based variant of our hybrid MAB rule.
$\phi_i$ denotes the latest contribution estimate of client~$i$;
$\sigma_i$ the number of times that client has been selected so far;
$t$ the current round index; $\epsilon$ the exploration probability;
$c$ the confidence weight; and $\tau$ the contribution floor that
triggers penalisation.
With probability~$\epsilon$ the server \emph{explores} by drawing $k$
clients uniformly without replacement, ensuring that every participant
is sampled infinitely often in expectation.  
Otherwise it \emph{exploits} by assigning each client the score
\begin{equation}
\label{eq:score}
  s_i \;=\;
  g_i
  \;+\;
  c\sqrt{\frac{\ln(t+1)}{\sigma_i+1}},
\end{equation}
where $t$ is the current round index, $\sigma_i$ counts previous
selections of client~$i$, and $c>0$ controls the confidence bonus.
The effective gain $g_i$ equals the latest contribution
$\phi_i$ if $\phi_i\ge\tau$ and $0$ otherwise, with
$\tau$ a small floor that penalises chronically under-performing
clients.
We fix the low-gain confidence scale at $0.1$, a conventional cap in $\epsilon$-greedy tuning that prevents uncertainty from overwhelming weak gains. A coarse grid $\{0.05,0.1,0.2\}$ gave nearly identical client rankings across datasets, with $0.1$ the most stable. 
The resulting non-negative score vector is normalised to a probability
distribution, and $k$ clients are drawn without replacement.

% \vspace{0.4em}
%---------------------------------------------------------------
% Algorithm : Hybrid Adaptive Client Selection
%---------------------------------------------------------------
\begin{algorithm}[ht]
\caption{Hybrid Adaptive Client Selection in FedOwen}
\label{alg:adaptive_selection}
\begin{algorithmic}[1]
\REQUIRE Contributions $\phi$; number to select $k$; exploration rate $\epsilon$; counters $\sigma$; round $t$; confidence weight $c$; contribution floor $\tau$
\ENSURE  Selected client index set $\mathcal K$
\IF{Uniform$(0,1)<\epsilon$}          % exploration
   \STATE $\mathcal K\gets$ random sample of $k$ distinct clients
\ELSE                                     % exploitation
   \STATE $s_i\gets 0$\; for all $i$
   \FOR{each client $i$}
      \STATE \textbf{if} $\phi_i<\tau$ \textbf{then} $g_i\gets 0$ \textbf{else} $g_i\gets\phi_i$
      \STATE $u_i\gets c\sqrt{\ln(t+1)/( \sigma_i + 1 )}$
      \IF{$\phi_i<\tau$} \STATE $u_i\gets 0.1\,u_i$ \ENDIF
      \STATE $s_i\gets g_i+u_i$
   \ENDFOR
   \STATE Shift $s_i\gets s_i-\min_j s_j$ so that $s_i\ge0$
   \STATE \textbf{if} $\sum_j s_j=0$ \textbf{then} set all $p_i\gets 1/n$
          \textbf{else} $p_i\gets s_i/\sum_j s_j$
   % \STATE $\mathcal K\gets$ weighted sample of $k$ distinct clients using probabilities $p$
   \STATE $\mathcal{K} \gets$ sample $k$ clients without replacement using weights $p$
\ENDIF
\STATE \textbf{return} $\mathcal K$
\end{algorithmic}
\end{algorithm}

\subsection{Contribution‐Aware Weighted Aggregation}
\label{subsec:weighted_agg}

After each valuation phase the server converts the contribution vector
$\boldsymbol{\phi}=(\phi_1,\dots,\phi_n)$ into a simplex‐valued
\emph{aggregation weight} vector
\begin{equation}
\alpha_i
\;=\;
\frac{\exp(\phi_i)}{\sum_{j=1}^{n}\exp(\phi_j)},
\qquad i = 1, \dots, n,
\end{equation}
so $\alpha_i\!\ge\!0$ and $\sum_i \alpha_i = 1$.  Each participating
client $i$ has trained a local model $W_i^{(t+1)}$ starting from the
previous global model $W^{(t)}$.  The new global model is obtained by a
FedAvg‐style weighted average of the \emph{full} local
weights:
\begin{equation}
W^{(t+1)}
\;=\;
\sum_{i=1}^{n} \alpha_i \, W_i^{(t+1)}.
\end{equation}
Because $\{\alpha_i\}$ form a convex combination, this update
automatically down-weights clients with low contribution scores and
leans toward those that were most helpful.

Figure \ref{fig:system_model} sketches the full round.  
Clients push their local models to the server; the server rates each one with fast Owen-sampling-plus-truncation, feeds those scores into an \(\epsilon\)-greedy bandit to pick the next roster, gives higher-scoring clients a bigger share when averaging, and finally broadcasts the refreshed global model back. The loop then repeats.

\section{Experimental Results}
\label{sec:exp}

\begin{table*}[!htb]
\centering
\caption{Final Model Accuracy for Different Approaches with Imbalance Factor = 0.01}
\resizebox{\textwidth}{!}{
\begin{tabular}{lccccccccccccccc}
\toprule
\textbf{Approach} & \multicolumn{3}{c}{\textbf{MedMNIST}} & \multicolumn{3}{c}{\textbf{MNIST}} & \multicolumn{3}{c}{\textbf{FashionMNIST}} & \multicolumn{3}{c}{\textbf{CIFAR-10}} & \multicolumn{3}{c}{\textbf{FEMNIST}} \\
\cmidrule(lr){2-4} \cmidrule(lr){5-7} \cmidrule(lr){8-10} \cmidrule(lr){11-13} \cmidrule(lr){14-16}
\textbf{Dirichlet $\alpha$} & \textbf{0.01} & \textbf{0.05} & \textbf{0.1} & \textbf{0.01} & \textbf{0.05} & \textbf{0.1} & \textbf{0.01} & \textbf{0.05} & \textbf{0.1} & \textbf{0.01} & \textbf{0.05} & \textbf{0.1} & \textbf{0.01} & \textbf{0.05} & \textbf{0.1} \\
\midrule
FedAvg & 15.42 & 23.33 & 33.33 & 77.60 & 83.74 & 90.71 & 59.96 & 65.59 & 72.97 & 21.48 & 23.13 & 28.81 & 19.44 & 20.41 & 20.82 \\
MC-Shapley \cite{castro2009polynomial} & 18.36 & 29.04 & 40.21 & 84.08 & \textbf{88.14} & 91.85 & 65.64 & 69.90 & 74.25 & 24.88 & 28.36 & 33.94 & 21.28 & 21.91 & 23.02 \\
GTG-Shapley \cite{liu2022gtg} & 15.31 & 26.00 & 37.24 & 79.22 & 84.60 & 90.61 & 60.90 & 66.59 & 73.55 & 22.99 & 24.42 & 30.17 & 20.07 & 20.62 & 22.08 \\
Data Banzhaf \cite{wang2023data} & 15.79 & 25.61 & 37.72 & 83.79 & 87.05 & 91.94 & 61.88 & 67.61 & 74.51 & 21.73 & 25.36 & 31.45 & 20.57 & 21.65 & 22.23 \\
WeightedSHAP \cite{kwon2022weightedshap} & 15.07 & 24.98 & 36.55 & 81.58 & 86.53 & 91.30 & 68.01 & 72.39 & 75.50 & 23.22 & 24.99 & 30.58 & 20.37 & 21.15 & 21.98 \\
ShapFed-WA \cite{tastan2024redefining} & 16.28 & 25.08 & 34.81 & 79.41 & 83.78 & 89.03 & 56.21 & 63.25 & 72.72 & 22.34 & 25.68 & 30.40 & 20.43 & 20.43 & 21.65 \\
FedOwen (ours) & \textbf{19.53} & \textbf{29.33} & \textbf{41.07} & \textbf{84.77} & 88.12 & \textbf{91.98} & \textbf{69.79} & \textbf{73.27} & \textbf{75.82} & \textbf{25.73} & \textbf{28.48} & \textbf{33.98} & \textbf{21.83} & \textbf{22.50} & \textbf{23.53} \\
\bottomrule
\end{tabular}
}
\label{tab:final_accuracies_001_un}
\end{table*}

\begin{table*}[!htb]
\centering
\caption{Final Model Accuracy for Different Approaches with Imbalance Factor = 0.05}
\resizebox{\textwidth}{!}{
\begin{tabular}{lccccccccccccccc}
\toprule
\textbf{Approach} & \multicolumn{3}{c}{\textbf{MedMNIST}} & \multicolumn{3}{c}{\textbf{MNIST}} & \multicolumn{3}{c}{\textbf{FashionMNIST}} & \multicolumn{3}{c}{\textbf{CIFAR-10}} & \multicolumn{3}{c}{\textbf{FEMNIST}} \\
\cmidrule(lr){2-4} \cmidrule(lr){5-7} \cmidrule(lr){8-10} \cmidrule(lr){11-13} \cmidrule(lr){14-16}
\textbf{Dirichlet $\alpha$} & \textbf{0.01} & \textbf{0.05} & \textbf{0.1} & \textbf{0.01} & \textbf{0.05} & \textbf{0.1} & \textbf{0.01} & \textbf{0.05} & \textbf{0.1} & \textbf{0.01} & \textbf{0.05} & \textbf{0.1} & \textbf{0.01} & \textbf{0.05} & \textbf{0.1} \\
\midrule
FedAvg & 19.52 & 27.93 & 37.33 & 81.12 & 86.96 & 93.66 & 62.67 & 69.09 & 77.00 & 22.08 & 25.90 & 32.50 & 22.79 & 23.91 & 24.87 \\
MC-Shapley \cite{castro2009polynomial} & 23.61 & 34.34 & 46.07 & 85.67 & 89.63 & 93.82 & 67.70 & 71.90 & 77.79 & 26.26 & 30.44 & \textbf{37.44} & 24.68 & 24.32 & 26.61 \\
GTG-Shapley \cite{liu2022gtg} & 20.27 & 31.42 & 42.21 & 82.60 & 87.16 & 93.47 & 63.55 & 69.71 & 77.29 & 23.24 & 27.58 & 34.42 & 23.65 & 25.43 & 25.62 \\
Data Banzhaf \cite{wang2023data} & 19.86 & 30.73 & 43.82 & 85.33 & 89.43 & 94.31 & 64.39 & 70.95 & 77.62 & 23.86 & 27.78 & 34.96 & 22.88 & 24.03 & 25.42 \\
WeightedSHAP \cite{kwon2022weightedshap} & 19.62 & 30.30 & 42.82 & 84.28 & 88.81 & 93.90 & 70.44 & 73.86 & 77.60 & 24.02 & 27.20 & 34.50 & 22.97 & 24.05 & 25.63 \\
ShapFed-WA \cite{tastan2024redefining} & 20.52 & 29.79 & 39.98 & 82.02 & 86.05 & 92.39 & 59.71 & 67.45 & 77.42 & 23.87 & 27.59 & 34.11 & 23.69 & 23.94 & 24.88 \\
FedOwen (ours) & \textbf{23.88} & \textbf{34.88} & \textbf{46.99} & \textbf{85.90} & \textbf{89.81} & \textbf{94.35} & \textbf{71.35} & \textbf{74.39} & \textbf{77.88} & \textbf{26.59} & \textbf{31.06} & 37.13 & \textbf{25.03} & \textbf{26.79} & \textbf{27.18} \\
\bottomrule
\end{tabular}
}
\label{tab:final_accuracies_005_un}
\end{table*}

These long-tailed configurations were intentionally chosen to verify robustness to class imbalance.

\subsection{Datasets and Settings}
\label{subsec:dataset}
We evaluate on MNIST \cite{lecun1998gradient}, MedMNIST \cite{yang2023medmnist}, FashionMNIST \cite{xiao2017fashion}, FEMNIST \cite{caldas2018leaf}, and CIFAR-10 \cite{krizhevsky2009learning}. To induce heterogeneity, we apply a long-tailed transform \cite{shang2022federated} (imbalance factors 0.01/0.05) and then a Dirichlet client split with $\alpha\in\{0.01,0.05,0.1\}$. The server holds a 1\% evaluation set isolated from training. Backbones are LeNet (all but CIFAR-10) and ResNet-18 (CIFAR-10). There are 100 clients; 10\% are sampled per round; training runs 100 rounds with one local epoch. Sensitivity sweeps vary $\epsilon$ and $Q$; other details are in the supplement.

For evaluation, we extract 1\% of the original dataset as an independent evaluation dataset, completely isolated from the training process. This independent set provides an unbiased measure of model performance and client contributions, avoiding the biases and privacy risks associated with validation-set evaluations. It ensures a fair and privacy-preserving contribution estimation, while effectively guiding global model improvements.

For MNIST, FashionMNIST, MedMNIST, and FEMNIST, we use LeNet as the model backbone. For CIFAR-10, we employ ResNet-18 due to the dataset’s increased complexity. Each experiment involves 100 clients, with 10\% of the clients (10 clients) selected per communication round. Training runs for 100 rounds, with each selected client performing one local epoch per selection.

In adaptive selection experiments, the $\epsilon$-value balancing exploration and exploitation is set to 0.1. We also perform a sensitivity analysis to examine the impact of varying $\epsilon$-values and the $Q$ parameter in Owen sampling. This ensures the robustness of the algorithm under different configurations.

In addition to FedAvg, we benchmark FedOwen against five prominent baselines: (1) \textit{MC-Shapley} \cite{castro2009polynomial}, estimating client contributions via random permutations; (2) \textit{Data Banzhaf} \cite{wang2023data}, evaluating contributions over random coalitions; (3) \textit{GTG-Shapley} \cite{liu2022gtg}, applying guided truncation sampling for efficient estimation; (4) \textit{WeightedShap} \cite{kwon2022weightedshap}, weighting contributions with a Beta-shaped distribution; and (5) \textit{ShapFed-WA} \cite{tastan2024redefining}, utilizing class-specific gradient-based Shapley values for aggregation. Further details on these baselines are provided in the supplementary material (\ref{app:baselines}).

All estimators in our study and our truncated Owen sampler, are executed within the \textbf{same} $\epsilon$‑greedy MAB scheduler and contribution‑weighted FedAvg aggregation (see Section~\ref{subsec:mab_selection}), so any performance difference is attributable solely to the valuation method.
Hence the only degree of freedom that differs across methods is the way they compute per‑client contributions~$\phi_i$.
This design isolates the estimator factor and rules out confounding from disparate selection or weighting policies.
\subsection{Metrics and Parameters}

We enforce budget parity by limiting every method to $N\times M$ calls to the utility $\nu(\cdot)$ per round, where $N$ is the number of clients and $M$ the sampling budget. MC-Shapley, GTG-Shapley, and Data Banzhaf run $M$ permutations; Owen sampling uses $Q$ inclusion levels with $M/Q$ permutations each, and any evaluations saved by early truncation are immediately recycled so the shared counter reaches exactly $N\times M$ calls. WeightedSHAP and ShapFed-WA are capped identically. All methods therefore share the $\mathcal{O}(NM)$ cost. We use $M=4$ and $Q=2$, so any accuracy gap reflects algorithmic quality rather than extra computation.

Experiments are conducted using PyTorch~2.3 on Ubuntu~20.04 with an Intel(R) Core(TM) i7-10700K CPU @ 3.80\,GHz and an NVIDIA GeForce RTX~3070 GPU. Random seeds are set to 10 different values for all experiments to ensure reproducibility. On our reference box, mean valuation time is $\approx 0.2$\,s per sampling round for $n=10$, $M=4$, $Q=2$; FedOwen does not add communication rounds (valuation is performed server-side).

\begin{figure*}[!ht]
    \centering
    \includegraphics[width=0.85\linewidth]{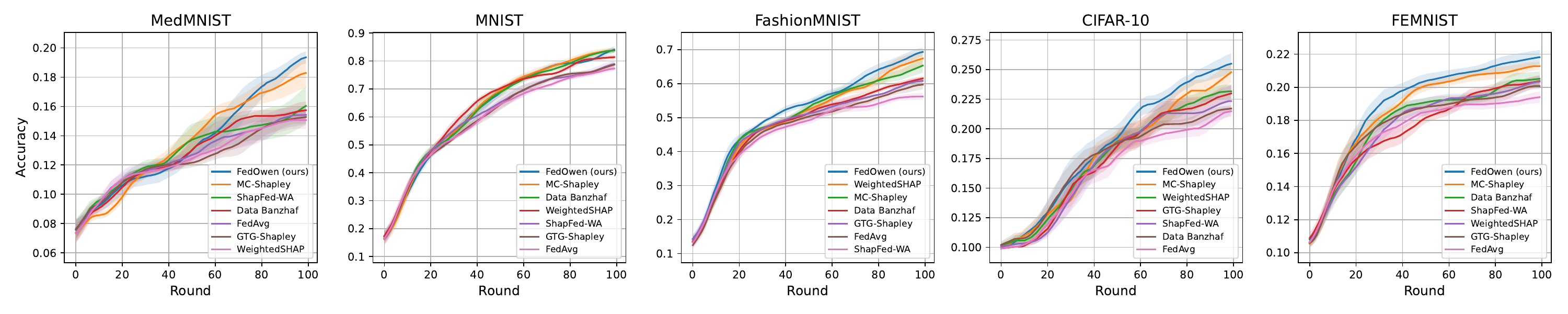}
    \vspace{-1.5em}
    \caption{Model performance over rounds with imbalance 0.01 and Dirichlet $\alpha$ = 0.01, averaged across seeds.}
    \label{fig:accuracy_un}
\end{figure*}

Figure~\ref{fig:accuracy_un} presents the model performance across datasets under an imbalance factor of 0.01 and Dirichlet heterogeneity with $\alpha$=0.01. Results for other parameter settings are provided in the supplementary material (\ref{app:addition}).
The results highlight the benefits of the proposed method in terms of higher final accuracies.

Tables \ref{tab:final_accuracies_001_un} and \ref{tab:final_accuracies_005_un} summarize performance across all settings. FedOwen improves final accuracy by up to 23\% over competing valuation estimators; in the two configurations where it falls just short, the gap is below half a percentage point, likely because these datasets are already near‑saturated. Taken together, the evidence underscores the strength of the proposed adaptive strategy.

\subsection{Sensitivity Analysis}
We study the effect of the inclusion levels $Q$ (Owen sampling) and the exploration rate $\epsilon$ (adaptive selection) under fixed heterogeneity ($\alpha=0.01$, imbalance $=0.01$).

\begin{figure*}[ht!]
    \centering
    \includegraphics[width=0.85\linewidth]{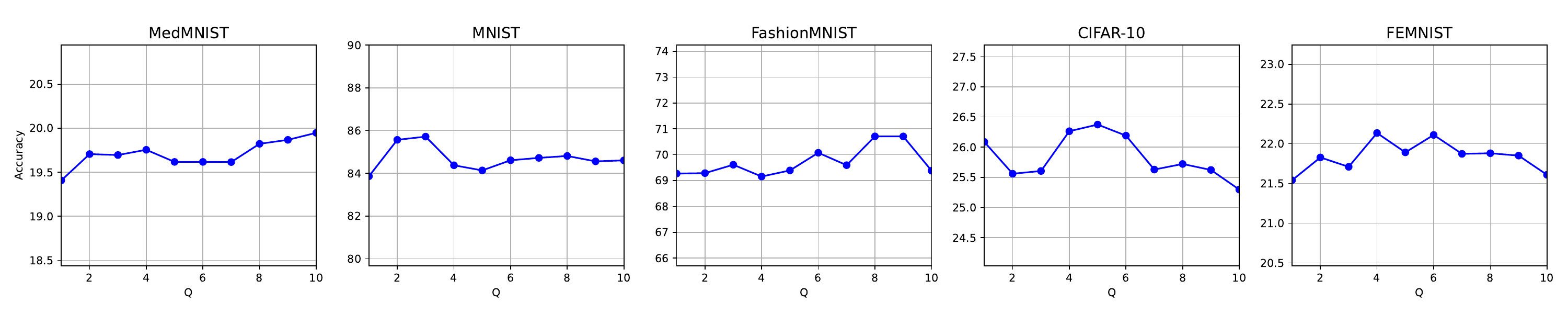}
    \vspace{-1.5em}
    \caption{Effect of inclusion levels (Q) on Final Accuracy}
    \label{fig:sens_q_un}
\end{figure*}

Figure~\ref{fig:sens_q_un} shows that increasing $Q$ yields modest accuracy gains but also higher computation; benefits taper quickly as $Q$ grows.

\begin{figure*}[ht!]
    \centering
    \includegraphics[width=0.85\linewidth]{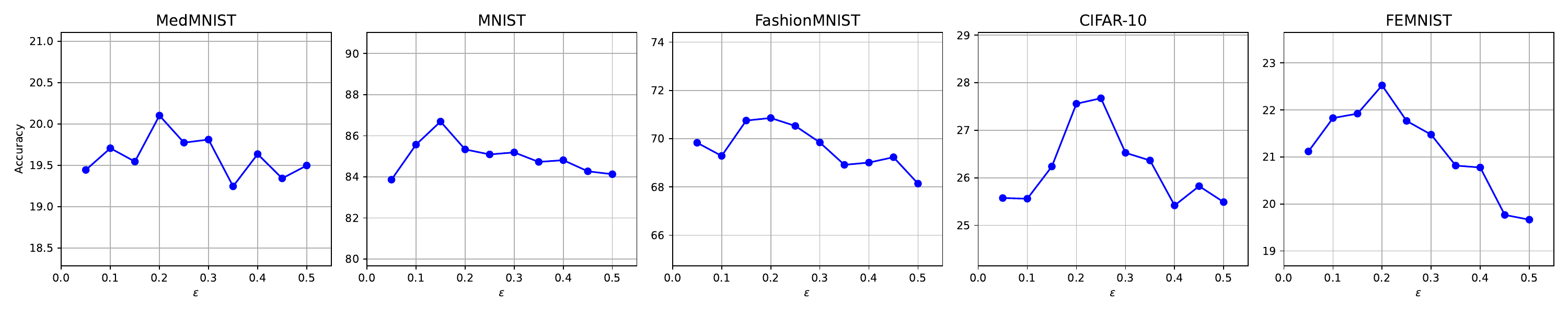}
    \vspace{-1.5em}
    \caption{Effect of $\epsilon$ on Final Accuracy}
    \label{fig:sens_eps_un}
\end{figure*}

Figure~\ref{fig:sens_eps_un} illustrates the exploration–exploitation trade-off. Larger $\epsilon$ improves diversity and helps avoid overfitting but can slightly reduce peak accuracy; smaller $\epsilon$ accelerates exploitation yet risks bias and overshooting (notably on FashionMNIST).

% To evaluate the impact of the $Q$ parameter in Owen sampling and the $\epsilon$-value in adaptive selection, we conduct sensitivity experiments under fixed heterogeneity ($\alpha=0.01$, imbalance factor = 0.01).

% Figure~\ref{fig:sens_q_un} shows that while increasing $Q$ can sometimes yield slightly better results, it also increases computational cost.  
% Beyond a certain point, gains in accuracy diminish relative to the added overhead.

% Figure~\ref{fig:sens_eps_un} illustrates the effect of different $\epsilon$-values on final accuracy.  
% Higher $\epsilon$ values (e.g., 0.1) introduce more exploration, which can help prevent overfitting but may slightly reduce maximum accuracy.  
% Conversely, smaller $\epsilon$ values promote faster exploitation but increase the risk of overshooting or bias, as observed in the FashionMNIST experiments.

Sensitivity analysis confirms that our adaptive algorithm is robust across a reasonable range of $\epsilon$ and $Q$ settings.

\begin{table*}[!htb]
\centering
\caption{Ablation Study on the Effect of Adaptive Client Selection in FedOwen for Two Imbalance Factors}
\resizebox{\textwidth}{!}{
\begin{tabular}{lccccccccccccccc}
\toprule
\textbf{FedOwen Variant} 
& \multicolumn{3}{c}{\textbf{MedMNIST}} 
& \multicolumn{3}{c}{\textbf{MNIST}} 
& \multicolumn{3}{c}{\textbf{FashionMNIST}} 
& \multicolumn{3}{c}{\textbf{CIFAR-10}} 
& \multicolumn{3}{c}{\textbf{FEMNIST}} \\
% & $\alpha$
\cmidrule(lr){2-4} \cmidrule(lr){5-7} \cmidrule(lr){8-10} \cmidrule(lr){11-13} \cmidrule(lr){14-16}
\textbf{Dirichlet $\alpha$} & \textbf{0.01} & \textbf{0.05} & \textbf{0.1} 
 & \textbf{0.01} & \textbf{0.05} & \textbf{0.1} 
 & \textbf{0.01} & \textbf{0.05} & \textbf{0.1} 
 & \textbf{0.01} & \textbf{0.05} & \textbf{0.1} 
 & \textbf{0.01} & \textbf{0.05} & \textbf{0.1} \\
\midrule
\textbf{w/o Adaptive (Imb. = 0.01)}    
& 18.31 & 27.38 & 38.1 
& 82.24 & 85.19 & 90.12 
& 67.81 & 70.41 & 73.25 
& 22.12 & 25.78 & 27.06 
& 20.14 & 20.48 & 21.14 \\
\textbf{with Adaptive (Imb. = 0.01)}   
& \textbf{19.53} & \textbf{29.33} & \textbf{41.07} 
& \textbf{84.77} & \textbf{88.12} & \textbf{91.98} 
& \textbf{69.79} & \textbf{73.27} & \textbf{75.82} 
& \textbf{25.73} & \textbf{28.48} & \textbf{33.98} 
& \textbf{21.83}    & \textbf{22.50}    & \textbf{23.53} \\
\midrule
\textbf{w/o Adaptive (Imb. = 0.05)}    
& 21.02 & 30.83 & 42.77
& 84.00 & 87.65 & 93.42
& 69.03 & 72.41 & 77.13
& 24.30 & 28.41 & 32.75
& 23.42 & 24.90 & 25.61 \\
\textbf{with Adaptive (Imb. = 0.05)}   
& \textbf{23.88} & \textbf{34.88} & \textbf{46.99} 
& \textbf{85.90} & \textbf{89.81} & \textbf{94.35} 
& \textbf{71.35} & \textbf{74.39} & \textbf{77.88} 
& \textbf{26.59} & \textbf{31.06} & \textbf{37.13} 
& \textbf{25.03}    & \textbf{26.79}    & \textbf{27.18} \\
\bottomrule
\end{tabular}
}
\label{tab:fedowen_ablation}
\end{table*}

\subsection{Ablation Study}
We next evaluate the impact of our adaptive client selection by turning off the MAB-based adaptation and relying solely on Owen sampling followed by random client selection with softmax‑normalized contribution weights.  

Table~\ref{tab:fedowen_ablation} reports the final model accuracy for both versions. Without adaptive selection, accuracy drops consistently across all datasets and heterogeneity settings. This confirms that using contribution history to guide client choice is essential to our gains.

\section{Conclusion}
We propose \textit{FedOwen}, which pairs Owen-sampled contribution estimation with an $\epsilon$-greedy client selector to balance exploration and exploitation. We deliberately include random picks in some rounds to preserve incentives and reduce overfitting, leading to more diverse participation and robust performance.

In experiments, \textit{FedOwen} achieves up to 23\% higher final accuracy than strong baselines, with the largest gains on long-tailed, highly non-IID splits. A sensitivity study shows stable performance across a wide range of $\epsilon$ and Owen-sampling settings.
% We propose FedOwen that integrates an $\epsilon$-greedy strategy with adaptive client contribution estimation in Federated Learning, balancing exploration and exploitation.  
% Randomized selections are deliberately incorporated in certain rounds to preserve fair incentives and prevent overfitting.  
% The hybrid selection strategy ensures both diversity and robust model performance.

% Experimental results show that FedOwen achieves up to 23\% higher final accuracy than the strongest baselines, with the largest gains on long‑tailed, highly non‑IID splits.  
% Sensitivity analysis confirms stable performance across a wide range of $\epsilon$ and Owen‑sampling parameters.
\paragraph{Limitations.}
(1) We rely on a fixed, server-held evaluation set; if it is unrepresentative, valuations may be biased.
(2) Our experiments scale to at most 100 clients; moving to thousands may require hierarchical sampling, stratified bandits, or more aggressive truncation.
(3) We assume limited client churn; fully dynamic join/leave behaviour remains open.
(4) Unbiasedness holds conditional on the utility $\nu$; robustness to adversarial or biased utilities (e.g., poisoned evaluation data) is orthogonal and left to future work (e.g., smoothing, robust caps, adversarial validation).
Future work also includes refining the bandit scheduler, adding differential-privacy safeguards, and better leveraging rare “maverick” clients.

\bibliography{references}

\begin{thebibliography}{62}
\providecommand{\natexlab}[1]{#1}
\providecommand{\url}[1]{\texttt{#1}}
\expandafter\ifx\csname urlstyle\endcsname\relax
  \providecommand{\doi}[1]{doi: #1}\else
  \providecommand{\doi}{doi: \begingroup \urlstyle{rm}\Url}\fi

\bibitem[Antunes et~al.(2022)Antunes, Andr{\'e}~da Costa, K{\"u}derle, Yari, and Eskofier]{antunes2022federated}
R.~S. Antunes, C.~Andr{\'e}~da Costa, A.~K{\"u}derle, I.~A. Yari, and B.~Eskofier.
\newblock Federated learning for healthcare: Systematic review and architecture proposal.
\newblock \emph{ACM Transactions on Intelligent Systems and Technology (TIST)}, 13\penalty0 (4):\penalty0 1--23, 2022.

\bibitem[Caldas et~al.(2018)Caldas, Duddu, Wu, Li, Kone{\v{c}}n{\`y}, McMahan, Smith, and Talwalkar]{caldas2018leaf}
S.~Caldas, S.~M.~K. Duddu, P.~Wu, T.~Li, J.~Kone{\v{c}}n{\`y}, H.~B. McMahan, V.~Smith, and A.~Talwalkar.
\newblock Leaf: A benchmark for federated settings.
\newblock \emph{arXiv preprint arXiv:1812.01097}, 2018.

\bibitem[Castro et~al.(2009)Castro, G{\'o}mez, and Tejada]{castro2009polynomial}
J.~Castro, D.~G{\'o}mez, and J.~Tejada.
\newblock Polynomial calculation of the shapley value based on sampling.
\newblock \emph{Computers \& operations research}, 36\penalty0 (5):\penalty0 1726--1730, 2009.

\bibitem[Chen and Xu(2024)]{chen2024federated}
K.~Chen and Z.~Xu.
\newblock Federated learning for data market: Shapley-ucb for seller selection and incentives.
\newblock \emph{arXiv preprint arXiv:2410.09107}, 2024.

\bibitem[Chen et~al.(2025)Chen, Zhang, Liu, Feng, and Yang]{chen2025pifl}
K.~Chen, J.~Zhang, X.~Liu, Z.~Feng, and X.~Yang.
\newblock $\pi$fl: Private, atomic, incentive mechanism for federated learning based on blockchain.
\newblock \emph{Blockchain: Research and Applications}, page 100271, 2025.

\bibitem[Chen et~al.(2024)Chen, Chen, Wang, and Chen]{chen2024space}
Y.-C. Chen, H.-W. Chen, S.-G. Wang, and M.-S. Chen.
\newblock Space: Single-round participant amalgamation for contribution evaluation in federated learning.
\newblock \emph{Advances in Neural Information Processing Systems}, 36, 2024.

\bibitem[Du et~al.(2025)Du, Wang, Li, Wang, and Fei]{du2025hfedcwa}
J.~Du, H.~Wang, J.~Li, K.~Wang, and R.~Fei.
\newblock Hfedcwa: heterogeneous federated learning algorithm based on contribution-weighted aggregation.
\newblock \emph{Applied Intelligence}, 55\penalty0 (2):\penalty0 1--16, 2025.

\bibitem[Ebron~Jr and Yang(2023)]{ebron2023fedtruth}
S.~C. Ebron~Jr and K.~Yang.
\newblock Fedtruth: Byzantine-robust and backdoor-resilient federated learning framework.
\newblock \emph{arXiv preprint arXiv:2311.10248}, 2023.

\bibitem[Fan et~al.(2022)Fan, Fang, Zhou, Pei, Friedlander, and Zhang]{fan2022fair}
Z.~Fan, H.~Fang, Z.~Zhou, J.~Pei, M.~P. Friedlander, and Y.~Zhang.
\newblock Fair and efficient contribution valuation for vertical federated learning.
\newblock \emph{arXiv preprint arXiv:2201.02658}, 2022.

\bibitem[Fu et~al.(2023)Fu, Zhang, Gao, Zhang, and Liu]{fu2023client}
L.~Fu, H.~Zhang, G.~Gao, M.~Zhang, and X.~Liu.
\newblock Client selection in federated learning: Principles, challenges, and opportunities.
\newblock \emph{IEEE Internet of Things Journal}, 2023.

\bibitem[Ghorbani and Zou(2019)]{ghorbani2019data}
A.~Ghorbani and J.~Zou.
\newblock Data shapley: Equitable valuation of data for machine learning.
\newblock In \emph{International conference on machine learning}, pages 2242--2251. PMLR, 2019.

\bibitem[Ghosh et~al.(2024)Ghosh, Basu, Huazhu, Yuan, Kanagavelu, Peng, Yong, Rick, and Qingsong]{ghosh2024don}
B.~Ghosh, D.~Basu, F.~Huazhu, W.~Yuan, R.~Kanagavelu, J.~J. Peng, L.~Yong, G.~S.~M. Rick, and W.~Qingsong.
\newblock Don't forget what i did?: Assessing client contributions in federated learning.
\newblock \emph{arXiv preprint arXiv:2403.07151}, 2024.

\bibitem[Guo et~al.(2024)Guo, Yao, Tian, Qi, Qi, Lin, and Dong]{guo2024contribution}
Q.~Guo, M.~Yao, Z.~Tian, S.~Qi, Y.~Qi, Y.~Lin, and J.~S. Dong.
\newblock Contribution evaluation of heterogeneous participants in federated learning via prototypical representations.
\newblock \emph{arXiv preprint arXiv:2407.02073}, 2024.

\bibitem[Huang et~al.(2021)Huang, Hong, Chen, and Roos]{huang2021shapley}
J.~Huang, C.~Hong, L.~Y. Chen, and S.~Roos.
\newblock Is shapley value fair? improving client selection for mavericks in federated learning.
\newblock \emph{arXiv preprint arXiv:2106.10734}, 2021.

\bibitem[Jalali and Hongsong(2025)]{jalali2025federated}
N.~A. Jalali and C.~Hongsong.
\newblock Federated learning incentivize with privacy-preserving for iot in edge computing in the context of b5g.
\newblock \emph{Cluster Computing}, 28\penalty0 (2):\penalty0 112, 2025.

\bibitem[Javaherian et~al.(2025)Javaherian, Turney, Chen, and Tzeng]{javaherian2025incentive}
S.~Javaherian, B.~Turney, L.~Chen, and N.-F. Tzeng.
\newblock Incentive-compatible federated learning with stackelberg game modeling.
\newblock \emph{arXiv preprint arXiv:2501.02662}, 2025.

\bibitem[Kolpaczki et~al.(2024)Kolpaczki, Bengs, Muschalik, and H{\"u}llermeier]{kolpaczki2024approximating}
P.~Kolpaczki, V.~Bengs, M.~Muschalik, and E.~H{\"u}llermeier.
\newblock Approximating the shapley value without marginal contributions.
\newblock In \emph{Proceedings of the AAAI Conference on Artificial Intelligence}, volume~38, pages 13246--13255, 2024.

\bibitem[Krizhevsky(2009)]{krizhevsky2009learning}
A.~Krizhevsky.
\newblock Learning multiple layers of features from tiny images.
\newblock Technical report, University of Toronto, 2009.

\bibitem[Kwon and Zou(2022)]{kwon2022weightedshap}
Y.~Kwon and J.~Y. Zou.
\newblock Weightedshap: analyzing and improving shapley based feature attributions.
\newblock \emph{Advances in Neural Information Processing Systems}, 35:\penalty0 34363--34376, 2022.

\bibitem[LeCun et~al.(1998)LeCun, Bottou, Bengio, and Haffner]{lecun1998gradient}
Y.~LeCun, L.~Bottou, Y.~Bengio, and P.~Haffner.
\newblock Gradient-based learning applied to document recognition.
\newblock \emph{Proceedings of the IEEE}, 86\penalty0 (11):\penalty0 2278--2324, 1998.

\bibitem[Li et~al.(2024)Li, Li, Zhou, and Yan]{li2024adafl}
Q.~Li, X.~Li, L.~Zhou, and X.~Yan.
\newblock Adafl: Adaptive client selection and dynamic contribution evaluation for efficient federated learning.
\newblock In \emph{ICASSP 2024-2024 IEEE International Conference on Acoustics, Speech and Signal Processing (ICASSP)}, pages 6645--6649. IEEE, 2024.

\bibitem[Li et~al.(2025)Li, Lyu, and Wen]{li2025optimal}
Q.~Li, S.~Lyu, and J.~Wen.
\newblock Optimal client selection of federated learning based on compressed sensing.
\newblock \emph{IEEE Transactions on Information Forensics and Security}, 2025.

\bibitem[Lin et~al.(2022)Lin, Xu, Liu, Li, Huang, and Shi]{lin2022contribution}
W.~Lin, Y.~Xu, B.~Liu, D.~Li, T.~Huang, and F.~Shi.
\newblock Contribution-based federated learning client selection.
\newblock \emph{International Journal of Intelligent Systems}, 37\penalty0 (10):\penalty0 7235--7260, 2022.

\bibitem[Liu et~al.(2025)Liu, Chang, Ye, Mumtaz, and H{\"a}m{\"a}l{\"a}inen]{liu2025game}
J.~Liu, Z.~Chang, C.~Ye, S.~Mumtaz, and T.~H{\"a}m{\"a}l{\"a}inen.
\newblock Game-theoretic power allocation and client selection for privacy-preserving federated learning in iomt.
\newblock \emph{IEEE Transactions on Communications}, 2025.

\bibitem[Liu et~al.(2024)Liu, Chang, Liu, Li, and Wang]{liu2024fairfed}
Y.~Liu, S.~Chang, Y.~Liu, B.~Li, and C.~Wang.
\newblock Fairfed: Improving fairness and efficiency of contribution evaluation in federated learning via cooperative shapley value.
\newblock In \emph{IEEE INFOCOM 2024-IEEE Conference on Computer Communications}, pages 621--630. IEEE, 2024.

\bibitem[Liu et~al.(2022)Liu, Chen, Yu, Liu, and Cui]{liu2022gtg}
Z.~Liu, Y.~Chen, H.~Yu, Y.~Liu, and L.~Cui.
\newblock Gtg-shapley: Efficient and accurate participant contribution evaluation in federated learning.
\newblock \emph{ACM Transactions on Intelligent Systems and Technology (TIST)}, 13\penalty0 (4):\penalty0 1--21, 2022.

\bibitem[Lv et~al.(2021)Lv, Zheng, Luo, Wu, Tang, Hua, Jia, and Lv]{lv2021data}
H.~Lv, Z.~Zheng, T.~Luo, F.~Wu, S.~Tang, L.~Hua, R.~Jia, and C.~Lv.
\newblock Data-free evaluation of user contributions in federated learning.
\newblock In \emph{2021 19th International Symposium on Modeling and Optimization in Mobile, Ad hoc, and Wireless Networks (WiOpt)}, pages 1--8. IEEE, 2021.

\bibitem[Mangla(2022)]{mangla2022application}
U.~Mangla.
\newblock Application of federated learning in telecommunications and edge computing.
\newblock In \emph{Federated Learning: A Comprehensive Overview of Methods and Applications}, pages 523--534. Springer, 2022.

\bibitem[Mitchell et~al.(2022)Mitchell, Cooper, Frank, and Holmes]{mitchell2022sampling}
R.~Mitchell, J.~Cooper, E.~Frank, and G.~Holmes.
\newblock Sampling permutations for shapley value estimation.
\newblock \emph{Journal of Machine Learning Research}, 23\penalty0 (43):\penalty0 1--46, 2022.

\bibitem[Muschalik et~al.(2024)Muschalik, Baniecki, Fumagalli, Kolpaczki, Hammer, and H{\"u}llermeier]{muschalik2024shapiq}
M.~Muschalik, H.~Baniecki, F.~Fumagalli, P.~Kolpaczki, B.~Hammer, and E.~H{\"u}llermeier.
\newblock shapiq: Shapley interactions for machine learning.
\newblock \emph{Advances in Neural Information Processing Systems}, 37:\penalty0 130324--130357, 2024.

\bibitem[Nagalapatti and Narayanam(2021)]{nagalapatti2021game}
L.~Nagalapatti and R.~Narayanam.
\newblock Game of gradients: Mitigating irrelevant clients in federated learning.
\newblock In \emph{Proceedings of the AAAI Conference on Artificial Intelligence}, volume~35, pages 9046--9054, 2021.

\bibitem[Okhrati and Lipani(2021)]{okhrati2021multilinear}
R.~Okhrati and A.~Lipani.
\newblock A multilinear sampling algorithm to estimate shapley values.
\newblock In \emph{2020 25th International Conference on Pattern Recognition (ICPR)}, pages 7992--7999. IEEE, 2021.

\bibitem[Owen(1972)]{owen1972multilinear}
G.~Owen.
\newblock Multilinear extensions of games.
\newblock \emph{Management Science}, 18\penalty0 (5-part-2):\penalty0 64--79, 1972.

\bibitem[Pan et~al.(2025)Pan, Li, Guan, Liang, Li, Wang, and Kou]{pan2025rfcsc}
Z.~Pan, Y.~Li, Z.~Guan, M.~Liang, A.~Li, J.~Wang, and F.~Kou.
\newblock Rfcsc: Communication efficient reinforcement federated learning with dynamic client selection and adaptive gradient compression.
\newblock \emph{Neurocomputing}, 612:\penalty0 128672, 2025.

\bibitem[Shang et~al.(2022)Shang, Lu, Huang, and Wang]{shang2022federated}
X.~Shang, Y.~Lu, G.~Huang, and H.~Wang.
\newblock Federated learning on heterogeneous and long-tailed data via classifier re-training with federated features.
\newblock \emph{arXiv preprint arXiv:2204.13399}, 2022.

\bibitem[Shi and Shen(2021)]{shi2021federated}
C.~Shi and C.~Shen.
\newblock Federated multi-armed bandits.
\newblock In \emph{Proceedings of the AAAI Conference on Artificial Intelligence}, volume~35, pages 9603--9611, 2021.

\bibitem[Shyn et~al.(2021)Shyn, Kim, and Kim]{shyn2021fedccea}
S.~K. Shyn, D.~Kim, and K.~Kim.
\newblock Fedccea: A practical approach of client contribution evaluation for federated learning.
\newblock \emph{arXiv preprint arXiv:2106.02310}, 2021.

\bibitem[Singhal et~al.(2024)Singhal, Pandey, and Popovski]{singhal2024greedy}
P.~Singhal, S.~R. Pandey, and P.~Popovski.
\newblock Greedy shapley client selection for communication-efficient federated learning.
\newblock \emph{IEEE Networking Letters}, 2024.

\bibitem[Song et~al.(2019)Song, Tong, and Wei]{song2019profit}
T.~Song, Y.~Tong, and S.~Wei.
\newblock Profit allocation for federated learning.
\newblock In \emph{2019 IEEE International Conference on Big Data (Big Data)}, pages 2577--2586. IEEE, 2019.

\bibitem[Sun et~al.(2023)Sun, Li, Zhang, Xiong, Liu, Liu, Qin, and Ren]{sun2023shapleyfl}
Q.~Sun, X.~Li, J.~Zhang, L.~Xiong, W.~Liu, J.~Liu, Z.~Qin, and K.~Ren.
\newblock Shapleyfl: Robust federated learning based on shapley value.
\newblock In \emph{Proceedings of the 29th ACM SIGKDD Conference on Knowledge Discovery and Data Mining}, pages 2096--2108, 2023.

\bibitem[Suzumura et~al.(2022)Suzumura, Zhou, Kawahara, Baracaldo, and Ludwig]{suzumura2022federated}
T.~Suzumura, Y.~Zhou, R.~Kawahara, N.~Baracaldo, and H.~Ludwig.
\newblock Federated learning for collaborative financial crimes detection.
\newblock In \emph{Federated learning: A comprehensive overview of methods and applications}, pages 455--466. Springer, 2022.

\bibitem[Tang et~al.(2021)Tang, Shao, Chen, Ye, Wu, and Xiao]{tang2021optimizing}
Z.~Tang, F.~Shao, L.~Chen, Y.~Ye, C.~Wu, and J.~Xiao.
\newblock Optimizing federated learning on non-iid data using local shapley value.
\newblock In \emph{Artificial Intelligence: First CAAI International Conference, CICAI 2021, Hangzhou, China, June 5--6, 2021, Proceedings, Part II 1}, pages 164--175. Springer, 2021.

\bibitem[Tastan et~al.(2024)Tastan, Fares, Aremu, Horvath, and Nandakumar]{tastan2024redefining}
N.~Tastan, S.~Fares, T.~Aremu, S.~Horvath, and K.~Nandakumar.
\newblock Redefining contributions: Shapley-driven federated learning.
\newblock \emph{arXiv preprint arXiv:2406.00569}, 2024.

\bibitem[Wang et~al.(2019)Wang, Dang, and Zhou]{wang2019measure}
G.~Wang, C.~X. Dang, and Z.~Zhou.
\newblock Measure contribution of participants in federated learning.
\newblock In \emph{2019 IEEE international conference on big data (Big Data)}, pages 2597--2604. IEEE, 2019.

\bibitem[Wang and Jia(2023)]{wang2023data}
J.~T. Wang and R.~Jia.
\newblock Data banzhaf: A robust data valuation framework for machine learning.
\newblock In \emph{International Conference on Artificial Intelligence and Statistics}, pages 6388--6421. PMLR, 2023.

\bibitem[Wang et~al.(2020)Wang, Rausch, Zhang, Jia, and Song]{wang2020principled}
T.~Wang, J.~Rausch, C.~Zhang, R.~Jia, and D.~Song.
\newblock A principled approach to data valuation for federated learning.
\newblock \emph{Federated Learning: Privacy and Incentive}, pages 153--167, 2020.

\bibitem[Wang et~al.(2024)Wang, Li, Luo, Li, Guo, and Wang]{wang2024fast}
Y.~Wang, K.~Li, Y.~Luo, G.~Li, Y.~Guo, and Z.~Wang.
\newblock Fast, robust and interpretable participant contribution estimation for federated learning.
\newblock In \emph{2024 IEEE 40th International Conference on Data Engineering (ICDE)}, pages 2298--2311. IEEE, 2024.

\bibitem[Winter(2002)]{winter2002shapley}
E.~Winter.
\newblock The shapley value.
\newblock \emph{Handbook of game theory with economic applications}, 3:\penalty0 2025--2054, 2002.

\bibitem[Wu et~al.(2023)Wu, Zhu, Zhang, Chen, Wang, and Cui]{wu2023fedab}
C.~Wu, Y.~Zhu, R.~Zhang, Y.~Chen, F.~Wang, and S.~Cui.
\newblock Fedab: Truthful federated learning with auction-based combinatorial multi-armed bandit.
\newblock \emph{IEEE Internet of Things Journal}, 10\penalty0 (17):\penalty0 15159--15170, 2023.

\bibitem[Xia et~al.(2020)Xia, Quek, Guo, Wen, Yang, and Zhu]{xia2020multi}
W.~Xia, T.~Q. Quek, K.~Guo, W.~Wen, H.~H. Yang, and H.~Zhu.
\newblock Multi-armed bandit-based client scheduling for federated learning.
\newblock \emph{IEEE Transactions on Wireless Communications}, 19\penalty0 (11):\penalty0 7108--7123, 2020.

\bibitem[Xiao et~al.(2017)Xiao, Rasul, and Vollgraf]{xiao2017fashion}
H.~Xiao, K.~Rasul, and R.~Vollgraf.
\newblock Fashion-mnist: a novel image dataset for benchmarking machine learning algorithms.
\newblock \emph{arXiv preprint arXiv:1708.07747}, 2017.

\bibitem[Xu et~al.(2021)Xu, Wang, Wang, and Yao]{xu2021fedcm}
J.~Xu, S.~Wang, L.~Wang, and A.~C.-C. Yao.
\newblock Fedcm: Federated learning with client-level momentum.
\newblock \emph{arXiv preprint arXiv:2106.10874}, 2021.

\bibitem[Yang et~al.(2022{\natexlab{a}})Yang, Liu, Sun, Li, and Li]{yang2022wtdp}
C.~Yang, J.~Liu, H.~Sun, T.~Li, and Z.~Li.
\newblock Wtdp-shapley: Efficient and effective incentive mechanism in federated learning for intelligent safety inspection.
\newblock \emph{IEEE Transactions on Big Data}, 2022{\natexlab{a}}.

\bibitem[Yang et~al.(2023)Yang, Shi, Wei, Liu, Zhao, Ke, Pfister, and Ni]{yang2023medmnist}
J.~Yang, R.~Shi, D.~Wei, Z.~Liu, L.~Zhao, B.~Ke, H.~Pfister, and B.~Ni.
\newblock Medmnist v2-a large-scale lightweight benchmark for 2d and 3d biomedical image classification.
\newblock \emph{Scientific Data}, 10\penalty0 (1):\penalty0 41, 2023.

\bibitem[Yang et~al.(2024{\natexlab{a}})Yang, Jarin, Buyukates, Avestimehr, and Markopoulou]{yang2024maverick}
M.~Yang, I.~Jarin, B.~Buyukates, S.~Avestimehr, and A.~Markopoulou.
\newblock Maverick-aware shapley valuation for client selection in federated learning.
\newblock \emph{arXiv preprint arXiv:2405.12590}, 2024{\natexlab{a}}.

\bibitem[Yang et~al.(2022{\natexlab{b}})Yang, Tan, Peng, Xiang, and Niu]{yang2022federated}
X.~Yang, W.~Tan, C.~Peng, S.~Xiang, and K.~Niu.
\newblock Federated learning incentive mechanism design via enhanced shapley value method.
\newblock \emph{Wireless Communications and Mobile Computing}, 2022\penalty0 (1):\penalty0 9690657, 2022{\natexlab{b}}.

\bibitem[Yang et~al.(2024{\natexlab{b}})Yang, Xiang, Peng, Tan, Wang, Liu, and Ding]{yang2024federated}
X.~Yang, S.~Xiang, C.~Peng, W.~Tan, Y.~Wang, H.~Liu, and H.~Ding.
\newblock Federated learning incentive mechanism with supervised fuzzy shapley value.
\newblock \emph{Axioms}, 13\penalty0 (4):\penalty0 254, 2024{\natexlab{b}}.

\bibitem[Yu et~al.(2020)Yu, Liu, Liu, Chen, Cong, Weng, Niyato, and Yang]{yu2020fairness}
H.~Yu, Z.~Liu, Y.~Liu, T.~Chen, M.~Cong, X.~Weng, D.~Niyato, and Q.~Yang.
\newblock A fairness-aware incentive scheme for federated learning.
\newblock In \emph{Proceedings of the AAAI/ACM Conference on AI, Ethics, and Society}, pages 393--399, 2020.

\bibitem[Yu et~al.(2025)Yu, Yang, Chen, Lan, Xing, and Yu]{yu2025fedgac}
Y.~Yu, X.~Yang, Z.~Chen, Y.~Lan, Z.~Xing, and D.~Yu.
\newblock Fedgac: optimizing generalization in personalized federated learning via adaptive initialization and strategic client selection.
\newblock \emph{Cluster Computing}, 28\penalty0 (2):\penalty0 120, 2025.

\bibitem[Zhang et~al.(2024)Zhang, Zhao, Ebron, and Yang]{zhang2024towards}
M.~Zhang, H.~Zhao, S.~Ebron, and K.~Yang.
\newblock Towards fair, robust and efficient client contribution evaluation in federated learning.
\newblock \emph{arXiv preprint arXiv:2402.04409}, 2024.

\bibitem[Zhang et~al.(2022)Zhang, Li, and Yang]{zhang2022data}
Z.~Zhang, X.~Li, and S.~Yang.
\newblock Data pricing in vertical federated learning.
\newblock In \emph{2022 IEEE/CIC International Conference on Communications in China (ICCC)}, pages 932--937. IEEE, 2022.

\bibitem[Zheng et~al.(2022)Zheng, Cao, and Yoshikawa]{zheng2022secure}
S.~Zheng, Y.~Cao, and M.~Yoshikawa.
\newblock Secure shapley value for cross-silo federated learning (technical report).
\newblock \emph{arXiv preprint arXiv:2209.04856}, 2022.

\end{thebibliography}
\clearpage                  % required when changing column mode
\onecolumn  

\appendix
\section{Long-Tailed Class Distribution}
\label{app:long}
\begin{figure*}[h!]
  \centering
  \includegraphics[width=\linewidth]{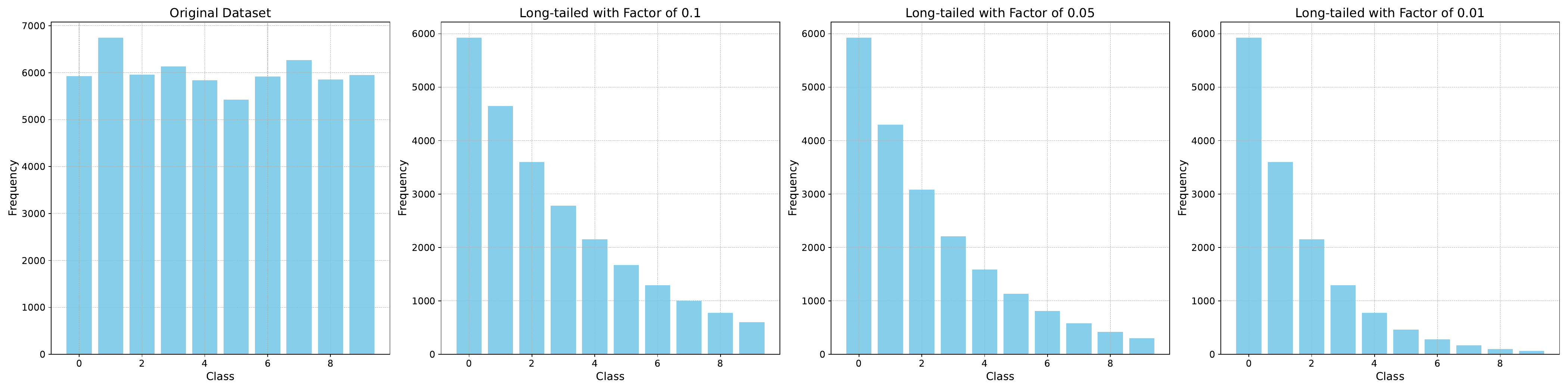}
  \caption{MNIST class counts before and after applying imbalance factors 0.1, 0.05, and 0.01.}
  \label{fig:mnist_long}
\end{figure*}

Figure~\ref{fig:mnist_long} illustrates how a long-tailed split changes the MNIST label histogram for three imbalance factors (0.1, 0.05, 0.01). MNIST is shown only as an example; the same transform is used for all datasets.

\vspace{8pt}

\vspace{8pt}

% ------------------------------------------------------------
\section{Additional Accuracy Curves}
\label{app:addition}
\begin{figure*}[!h]
  \centering
  \includegraphics[width=\linewidth]{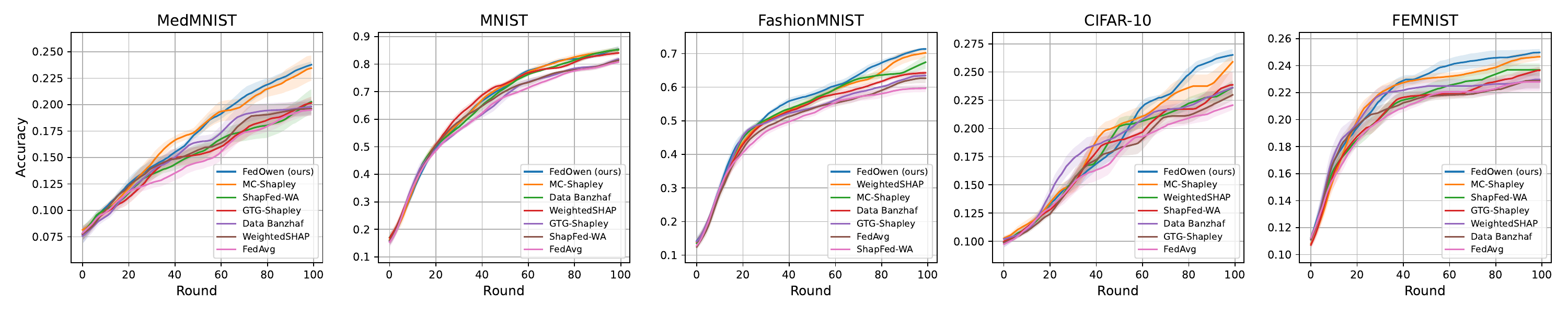}
  \caption{Imbalance 0.05, Dirichlet $\alpha=0.01$.}
  \label{fig:acc_alpha001_imb005}
\end{figure*}

\begin{figure*}[!h]
  \centering
  \includegraphics[width=\linewidth]{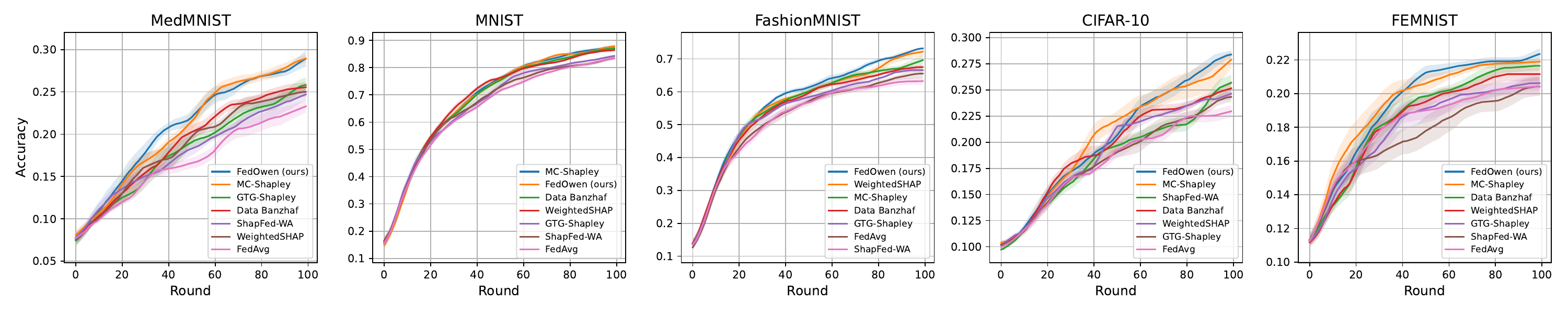}
  \caption{Imbalance 0.01, Dirichlet $\alpha=0.05$.}
  \label{fig:acc_alpha005_imb001}
\end{figure*}

\begin{figure*}[!h]
  \centering
  \includegraphics[width=\linewidth]{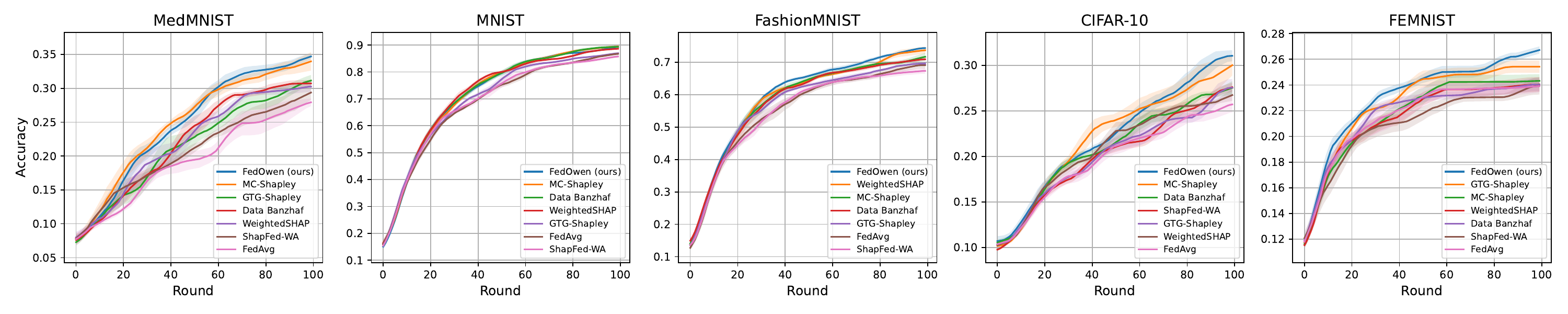}
  \caption{Imbalance 0.05, Dirichlet $\alpha=0.05$.}
  \label{fig:acc_alpha005_imb005}
\end{figure*}

\begin{figure*}[!h]
  \centering
  \includegraphics[width=\linewidth]{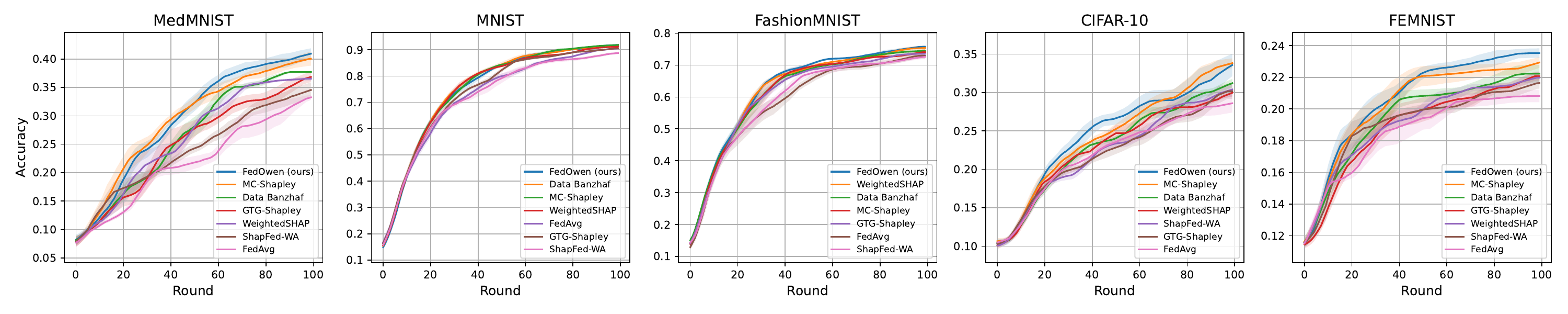}
  \caption{Imbalance 0.01, Dirichlet $\alpha=0.1$.}
  \label{fig:acc_alpha010_imb001}
\end{figure*}

\begin{figure*}[!h]
  \centering
  \includegraphics[width=\linewidth]{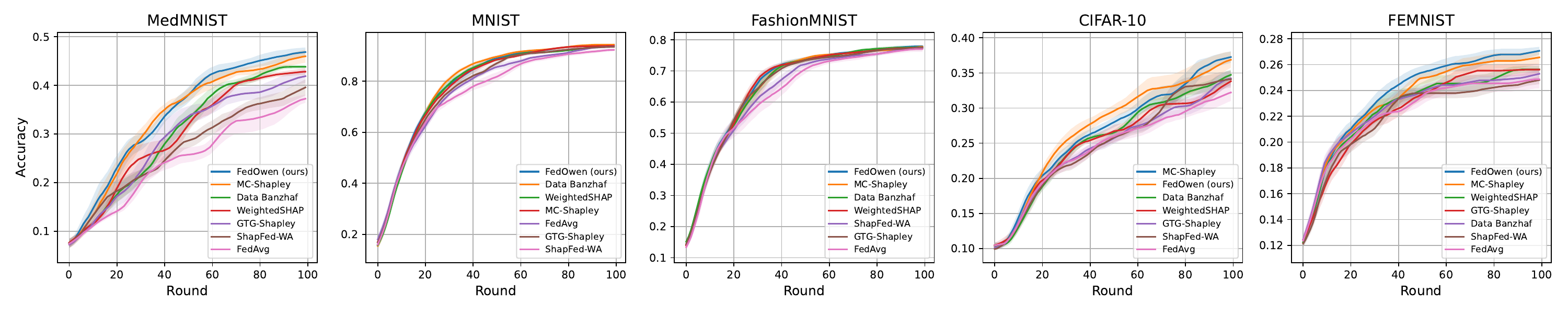}
  \caption{Imbalance 0.05, Dirichlet $\alpha=0.1$.}
  \label{fig:acc_alpha010_imb005}
\end{figure*}

Figures~\ref{fig:acc_alpha001_imb005}–\ref{fig:acc_alpha010_imb005}
report accuracy for the remaining pairs of Dirichlet $\alpha$ and imbalance factor
($\alpha\in\{0.01,0.05,0.1\}$; imbalance $\in\{0.01,0.05\}$).
All plots use the same layout as in the main paper. Each curve is averaged over three random seeds. \\
Across all settings, lower $\alpha$ or stronger imbalance slows
convergence, but the ranking of methods is consistent with the main paper.

\section{Baseline Algorithms}\label{sec:baselines}
\label{app:baselines}
We give pseudocode for each baseline estimator used in Section~\ref{sec:exp} of the main paper. All baselines are executed under the same
$N{\times}M$ evaluation budget and the shared $\epsilon$-greedy
scheduler described in Section~\ref{subsec:dataset}.

% ---------- MC-Shapley ----------
\paragraph{1.~MC-Shapley \cite{castro2009polynomial}.}
Randomly samples full permutations and averages marginal gains.
\begin{algorithm}[H]
  \caption{MC-Shapley approximation~\cite{castro2009polynomial}}
  \label{alg:mc_shapley}
  \begin{algorithmic}[1]
  \REQUIRE Client set $N$, samples $M$
  \STATE $\phi_i\gets0$ \textbf{for all} $i\in N$
  \FOR{$m=1$ \textbf{to} $M$}
     \STATE Draw random permutation $\pi$, set $S\gets\varnothing$
     \FOR{$i$ \textbf{in} $\pi$}
        \STATE $\phi_i \mathrel{+}= v(S\cup\{i\})-v(S)$
        \STATE $S \gets S\cup\{i\}$
     \ENDFOR
  \ENDFOR
  \STATE \textbf{return} $\phi_i/M$
  \end{algorithmic}
\end{algorithm}

% ---------- Data Banzhaf ----------
\paragraph{2.~Data Banzhaf \cite{wang2023data}.}
Estimates Banzhaf values by sampling random subsets instead of permutations.
\begin{algorithm}[H]
  \caption{Monte-Carlo Banzhaf approximation~\cite{wang2023data}}
  \label{alg:mc_banzhaf}
  \begin{algorithmic}[1]
  \REQUIRE Client set $N$, samples $M$
  \STATE $\beta_i\gets0$ \textbf{for all} $i\in N$
  \FOR{$m=1$ \textbf{to} $M$}
     \STATE Draw random subset $S\subseteq N$
     \FOR{$i\in N\setminus S$}
        \STATE $\beta_i \mathrel{+}= v(S\cup\{i\})-v(S)$
     \ENDFOR
  \ENDFOR
  \STATE \textbf{return} $\beta_i/M$
  \end{algorithmic}
\end{algorithm}

% ---------- GTG-Shapley ----------
\paragraph{3.~GTG-Shapley \cite{liu2022gtg}.}
Guided truncation skips marginal gains below a threshold $\epsilon$.
\begin{algorithm}[H]
  \caption{GTG-Shapley approximation~\cite{liu2022gtg}}
  \label{alg:gtg_shapley}
  \begin{algorithmic}[1]
  \REQUIRE Clients $N$, threshold $\epsilon$, samples $M$
  \STATE $\phi_i\gets0$, compute $v(N)$
  \FOR{$m=1$ \textbf{to} $M$}
     \STATE Draw permutation $\pi$, set $S\gets\varnothing$, $v_{\text{prev}}\gets0$
     \FOR{$i$ \textbf{in} $\pi$}
        \IF{$|v(N)-v_{\text{prev}}|\ge\epsilon$}
            \STATE $v_{\text{curr}}\gets v(S\cup\{i\})$
        \ELSE
            \STATE $v_{\text{curr}}\gets v_{\text{prev}}$
        \ENDIF
        \STATE $\phi_i \mathrel{+}= v_{\text{curr}}-v_{\text{prev}}$
        \STATE $v_{\text{prev}}\gets v_{\text{curr}}$, $S\gets S\cup\{i\}$
     \ENDFOR
  \ENDFOR
  \STATE \textbf{return} $\phi_i/M$
  \end{algorithmic}
\end{algorithm}

% ---------- WeightedSHAP ----------
\paragraph{4.~WeightedSHAP (Beta-shaped) \cite{kwon2022weightedshap}.}
Applies Beta weights to favour early or late coalition positions.
\begin{algorithm}[H]
  \caption{WeightedSHAP approximation~\cite{kwon2022weightedshap}}
  \label{alg:weightedshap}
  \begin{algorithmic}[1]
  \REQUIRE Clients $N$, Beta parameters $\alpha,\beta$, samples $M$
  \STATE Pre-compute Beta weights $w_j$ for $j=1,\dots,|N|$, set $\phi_i\gets0$
  \FOR{$m=1$ \textbf{to} $M$}
     \STATE Draw permutation $\pi$, set $S\gets\varnothing$
     \FOR{$j=1$ \textbf{to} $|N|$}
        \STATE $i\gets\pi[j]$, $\Delta\gets v(S\cup\{i\})-v(S)$
        \STATE $\phi_i \mathrel{+}= w_j\,\Delta$, $S\gets S\cup\{i\}$
     \ENDFOR
  \ENDFOR
  \STATE \textbf{return} $\phi_i/M$
  \end{algorithmic}
\end{algorithm}

% ---------- ShapFed-WA ----------
\paragraph{5.~ShapFed-WA \cite{tastan2024redefining}.}
Uses per-class Shapley values from last-layer gradients to weight aggregation.
\begin{algorithm}[H]
  \caption{ShapFed-WA aggregation~\cite{tastan2024redefining}}
  \label{alg:shapfed_wa}
  \begin{algorithmic}[1]
  \REQUIRE Class-specific gradients $\hat{w}_i$
  \STATE Compute contribution scores $\gamma_i$ via cosine similarity
  \STATE Normalise: $\tilde{\gamma}_i = \gamma_i / \sum_j \gamma_j$
  \STATE \textbf{return} $w_\text{global}=\sum_i \tilde{\gamma}_i\,\hat{w}_i$
  \end{algorithmic}
\end{algorithm}

% \bibliography{references}

\end{document}